# Giving life to robotic skins

Ahmad Rafsanjani[1a], Fergal B. Coulter[2b], André R. Studart[2c]

[1] Center for Soft Robotics, SDU Biorobotics, University of Southern Denmark, Odense, Denmark
[2] Complex Materials, Department of Materials, ETH Zurich, Zurich, Switzerland

Correspondence:
[a] ahra@mmmi.sdu.dk
[b] fergal.coulter@mat.ethz.ch
[c] andre.studart@mat.ethz.ch

**Abstract**

The skin of humanoid robots often lacks human tactility and the inherent self-repair capability of biological tissues. Recently, researchers have grown a living, self-healing skin on a robot finger by subsequent culturing of human dermal and epidermal cells. Here, we highlight the significance of this study alongside challenges toward developing biohybrid robots equipped with sensate and adaptive living robotic skins.

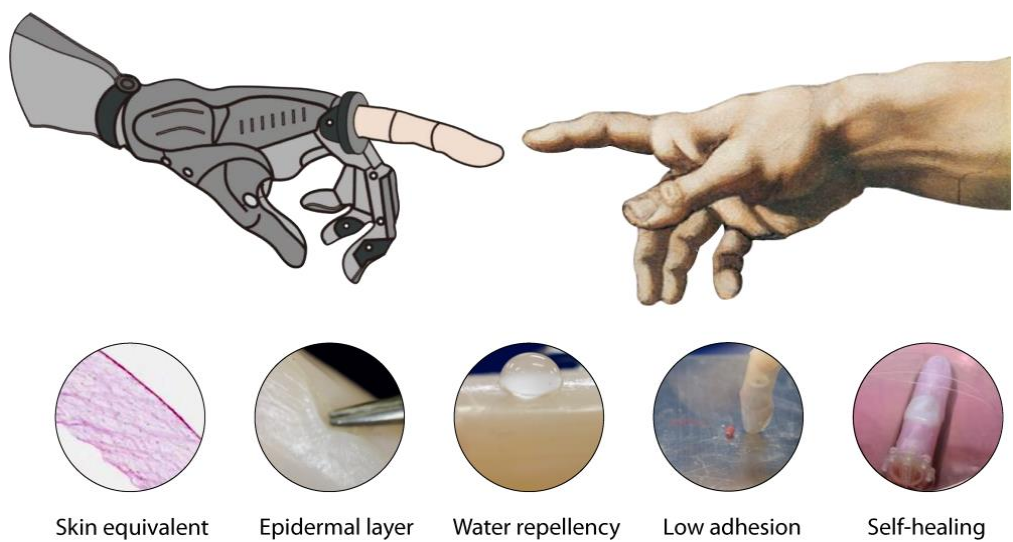

**Figure 1**. Living skin equivalent grown on a robotic finger featuring dermal and epidermal layers with water repellency, low adhesion, and self-healing characteristics.

Interactions between humanoid robots with humans and the environment has been focused almost exclusively on the face and voice, overlooking the importance of the skin – the largest organ in our body. Conversely, touch communicates distinct emotions among humans, such as anger, fear, disgust, love, gratitude, and sympathy [1]. Our skin is an active sensory organ, a socially expressive means, a permeable regulatory filter, and a self-healing protective layer [2]. In contrast, the skin of existing humanoid robots is a passive layer, the sole function of which is shielding the robot's internal structure from the external world. Robotics has taken tremendous leaps in generating extremely complex human gaits, as seen in the latest Atlas robot (Boston Dynamics) that can parkour with jumps and vaults like a real pro. However, the rigid insensate skin of existing humanoid robots is remarkably limited when it comes to interactions with humans or adapting to dynamic environments. In recent years, the underexplored world of robotic skins has attracted researchers from many disciplines to augment the interaction affordance of robots.

Research on robotic skins – covering fingertips to whole-body of robots – has been primarily focused on integrating electronic skins, known as *E-skins*, onto the external surface of robots to give them a sense of touch [3]. As an illustrative example, researchers developed a multilayered electronic dermis on a robot's fingertip made up of conductive and piezoresistive textiles encased in silicone rubber. Such an artificial skin can transform tactile information from object grasping into a neuromorphic signal and pass it to the peripheral nerves of an amputee to elicit sensory perceptions of touch and pain [4]. Unlike E-skins where electrons are responsible for sensing and transmission of tactile information, human skin sensors rely on ionic current. Ionic skins bring robotic skin equivalents closer to their biological counterparts and, in the simplest form, can be constructed by sandwiching a stretchable dielectric layer between two stretchable ionically conductive layers usually made of salt-containing hydrogels [5]. The assembly of charged poly(acrylic acid) and neutral polyacrylamide hydrogels resulted in a piezoionic sensor skin with a built-in potential difference that operates by pressure-driven ion flux and can find application in neuro-prosthetic robots [6]. Although essential in several applications, upscaling these and other tactile sensing technologies to the whole-body area is even more challenging, since it involves the organization and calibration of several spatially distributed discrete sensors and big data processing. Roboticists demonstrated the feasibility of covering almost the entire surface of a human-sized humanoid robot with more than a thousand self-calibrated multimodal sensing tiles, enabling the spatial perception of temperature, pressure, acceleration, and proximity [7].

Besides the sense of touch, self-healing and damage resilience have also been features of major interest in the development of soft materials and artificial skins for robotic applications [8]. The idea has been to imbue electronic skins with the ability to restore sensing functionalities by using materials that can re-establish broken interconnects or fill up cracks after damage. Such self-healing capabilities have been achieved in electronic skins using, for example, liquid metals as fluid conductive elements and dynamic supramolecular polymers and hydrogels as structural materials. In spite of these remarkable feats, self-healing artificial materials still suffer from excessive creep, dependence on external stimulus for healing and tendency to deteriorate with the number of healing events.

While research on robotic skins previously concentrated primarily on reproducing perceptual functions, giving robots a lifelike appearance, and generating a humanlike tactile feeling with self-repair capabilities is imperative for their broader dissemination and acceptance by humans in the medical care, nursing centers, and service industry. Recently, Takeuchi and coworkers developed a living human skin equivalent grown on a robot finger [9]. This skin exhibits a self-healing ability and can bear the deformation induced by the joint motions of the finger. A novelty of this research lies in the fabrication of an *in vitro* cell culture that uniformly covers a three-dimensional three-joint finger, demonstrating the implementation of a human-like skin on a fundamental building block of a humanoid robot hand. Living fibroblasts from connective tissue of human skin are used to build a dermal layer, which is covered with an epidermal layer containing living keratinocytes responsible for the barrier function of the human skin. Emphasis is placed on the fixation of the newly grown bilayer skin to the substrate, where anchor points were created to prevent slippage or incorrect movement of the coating. To manufacture the skin, the bare robotic finger was first inserted into a cylindrical rubber mold, with a small gap between the two. A collagen solution containing living fibroblasts was then poured into the mold and incubated. After three days, a mature dermis formed on the irregular three-dimensional surface of the finger that featured an anchor fixation at its root to minimize tissue shrinkage. Finally, the epidermal layer was grown atop the dermis tissue by seeding and culturing living keratinocytes from different sides of the mold for another two weeks to complete the construction of the skin equivalent. Histological analysis on frozen sections of stained skin equivalent proved a seamless adhesion of epidermis and dermis tissues and the formation of uniform layers.

The skin equivalent consisting of dermal and epidermal layers showed a significantly lower capacitance and a higher electrical resistance compared to the control sample containing only the dermis layer. This confirmed the effectiveness of the epidermis as an electrical barrier for the robot skin. Water retention tests demonstrated that the presence of the epidermis tissue makes the skin equivalent almost impermeable compared to the pure dermis equivalent. Moreover, the epidermis layer is sufficiently water repellent and non-adherent to form water droplets on its surface and to efficiently handle electrostatically charged objects (e.g., polystyrene foam beads). In addition to the epidermis, the performance of the dermal layer as the load-bearing component of the fabricated skin was also evaluated. Mechanical characterization of the dermis equivalent revealed a strain-hardening regime akin to human skin. However, the measured tensile strength of the artificial living skin is orders of magnitude lower than the values reported for the biological counterpart. While this low mechanical strength remains an open challenge, the skin showed cell-mediated self-repair capabilities and was sufficiently robust to withstand the stresses developed during motion of the finger. The self-healing capability of the living robotic skin is demonstrated by applying an acellular collagen sheet to an intentional wound on the dorsal dermis of the robotic finger. The grafted collagen sheet successfully sealed the wound and seamlessly adhered to the dermis equivalent after one week.

In the past, we have seen many examples of humanoid robots with skin equivalents that are fabricated using artificial polymers such as silicone rubber or foamed latex. While humanoids can be exquisitely painted, sculpted, and accented with make-up to improve their human-likeness, very often these robots fall into the category of not looking quite human enough to be convincing, or worse – they fall into the so-called *uncanny valley* - whereby people are left unsettled by the perceivable but subtle differences in the visible versus expected movement. Human-like resemblance can be improved by modifying the extensibility and compliance of the polymers themselves. Techniques used in theatrical prosthetics such as chain extenders or *deadeners* can modify the movement of the elastomer skin, improving on how wrinkling or folding occurs due to movement. Nonetheless, this is no substitute for the real thing, and these new conformal cell culturing techniques developed by Takeuchi and coworkers should facilitate an improved experience for future interactions with humanoid robots.

Some caveats regarding the concept of creating and sustaining a true *living skin* over a robot include a need for blood vessels and capillaries to carry artificial blood containing oxygen and nutrients required to maintain the cells. Alongside this, a distribution network should exist for stem cells to replace and remove apoptotic skin cells, a particularly challenging task given the delicate nature of undifferentiated cells. Integrating a sensor network into the dermis equivalent, potentially based on ionic currents, will also be required to add perception to the repertoire of living robotic skin functions. Finally, upscaling from a robot finger to a human-sized humanoid robot cannot be achieved with current traditional casting methods and requires recruiting digital manufacturing techniques such as multiaxial, multi-material, and multifunctional bioprinting of complex, soft materials [10]. Despite all these shortcomings, Takeuchi and coworkers showcased that giving life to robotic skins can enable the realization of expressive, protective, and self-healing biohybrid robots.

**Acknowledgment**
A.R was supported by the Villum Young Investigator grant 37499 and the Danish Council for Independent Research (DFF) through the DFF Sapere Aude grant 1051-00075B. A.R.S. and F.B.C. thank the financial support from the Swiss National Science Foundation (grant 200020_204614) and from the Strategic Focus Area Advanced Manufacturing (SFA-AM) of the Swiss ETH domain, as part of the Manufhaptics project.